\documentclass[runningheads]{llncs}

\usepackage{hyperref}
\usepackage{graphicx}
\usepackage{textcomp}
\usepackage{bbding}
\usepackage{xcolor}
\usepackage[export]{adjustbox}
\usepackage{amsmath}
\usepackage{amssymb}
\usepackage{mathtools}
\usepackage{multirow}
\usepackage{pgf}
\usepackage{tikz}
\usetikzlibrary{arrows, automata, shapes, petri, positioning, calc}
\usepackage[font=scriptsize,labelfont=bf]{caption}

\hypersetup{
	colorlinks=true,
	linkcolor={red!50!black},
	citecolor={blue!60!black},
	urlcolor={blue!80!black},
	pdfauthor={Pegoraro Marco, Bakullari Bianka, Uysal Merih Seran, van der Aalst Wil M.P.},
	pdftitle={{Probability Estimation of Uncertain Process Trace Realizations}},
	pdfsubject={Process Mining over Uncertain Data},
	pdfkeywords={Process Mining, Uncertain Data, Partial Order},
	pdfproducer={LaTeX},
	pdfcreator={pdfLaTeX},
	bookmarksopen=true
}

\begin{document}

\title{Probability Estimation of Uncertain\\Process Trace Realizations\thanks{We thank the Alexander von Humboldt (AvH) Stiftung for supporting our research interactions.}}

\author{Marco Pegoraro\,\Envelope\,\orcidID{0000-0002-8997-7517} \and Bianka Bakullari\orcidID{0000-0003-2680-0826} \and Merih Seran Uysal\orcidID{0000-0003-1115-6601} \and Wil M.P. van der Aalst\orcidID{0000-0002-0955-6940}}

\authorrunning{Pegoraro et al.}

\institute{Process and Data Science Group (PADS) \\ Department of Computer Science, RWTH Aachen University, Aachen, Germany
	\email{\{pegoraro, bianka.bakullari, uysal, wvdaalst\}@pads.rwth-aachen.de}\\
	\url{http://www.pads.rwth-aachen.de/}}

\maketitle

\begin{abstract}
Process mining is a scientific discipline that analyzes event data, often collected in databases called event logs. Recently, \emph{uncertain event logs} have become of interest, which contain non-deterministic and stochastic event attributes that may represent many possible real-life scenarios. In this paper, we present a method to reliably estimate the probability of each of such scenarios, allowing their analysis. Experiments show that the probabilities calculated with our method closely match the true chances of occurrence of specific outcomes, enabling more trustworthy analyses on uncertain data.

\end{abstract}

\section{Introduction}\label{sec:introduction}
Process mining is a discipline that focuses on extracting insights about processes in a data-driven manner. For instance, on the basis of the recorded information on historical process executions, process mining allows to automatically extract a model of the behavior of process instances, or to measure the compliance of the process data with a prescribed normative model of the process. In process mining, the central focus is on the \emph{event log}, a collection of data that tracks past process instances. Every activity performed in a process is recorded in the event log, together with information such as the corresponding process case and the timestamp of the activity, in a sequence of events called a \emph{trace}.

Recently, research on novel forms of event data have garnered the attention of the scientific community. Among these there are \emph{uncertain event logs}, which contain data affected by imprecision~\cite{pegoraro2019mining}. This data contains meta-information describing the nature and entity of the uncertainty. Such meta-information can be obtained from the inherent precision with which the data has been recorded (e.g., timestamps only indicating the date have a possible ``true value'' range of 24 hours), from the precision of the tools involved in supporting the process (e.g., the absolute error of sensors), or from the domain knowledge provided by a process expert. An uncertain trace corresponds to multiple possible real-life scenarios, each of which might have very diverse implications on features of cases such as compliance to a model. It is then important to be able to assess the risk of occurrence of specific outcomes of uncertain traces, which enables to estimate the impact of such traces on indicators such as cost and conformance.

In this paper, we present a method to obtain a complete probability distribution over the possible instantiations of uncertain attributes in a trace. As a possible example of application, we frame our results in the context of conformance checking, and show the impact of assessing probability estimates for uncertain traces on insights about the compliance of an uncertain trace to a process model. 
We validate our method with experiments based on a Monte Carlo simulation, which shows that the probability estimates are reliable and reflect the true chances of occurrence of a specific outcome.

The remainder of the paper is structured as follows. Section~\ref{sec:rel} examines relevant related work. Section~\ref{sec:example} illustrates a motivating running example for our technique. Section~\ref{sec:prelim} presents preliminary definitions of different types of uncertainty in process mining. Section~\ref{sec:meth} illustrates a method for computing probabilities of realizations for uncertain process traces. Section~\ref{sec:exp} validates our method through experimental results. Finally, Section~\ref{sec:conc} concludes the paper.

\section{Related Work}\label{sec:rel}
The analysis of uncertain data in process mining is a very recent research direction. The specific formulation and definition of uncertain data utilized in this paper has been introduced in 2019~\cite{pegoraro2019mining}, in the context of an analysis approach consisting in computing bounds for the conformance score of uncertain traces through alignments~\cite{van2017aligning}. Subsequently, that work has been extended with an inductive mining approach for process discovery over uncertainty~\cite{pegoraro2019discovering} and a taxonomy of different types of uncertain data, with their characteristics~\cite{DBLP:journals/informationsystems/PegoraroUA21}.

Uncertain data, as formulated in our present and previous work, is closely related to a considerably more studied data anomaly in process mining: partially ordered event data. In fact, uncertain data as described here is a generalization of partially ordered traces. Lu et al.~\cite{lu2014conformance} proposed a conformance checking approach based on alignments to measure conformance of partially ordered traces. More recently, Van der Aa et al.~\cite{van2020partial} illustrated a method for inferring a linear extension, i.e., a compliant total order, of events in partially ordered traces, based on examples of correct orderings extracted from other traces in the log. Busany et al.~\cite{busany2020interval} estimated probabilities for partially ordered events in IoT event streams.

An associated topic, which draws from disciplines such as pattern and sequence mining and is antithetical to the analysis of partially ordered data, is the inference of partial orders from fully sequential data as a way to model its behavior. This goes under the name of \emph{episode mining}, which can be performed with many techniques both on batched data and with online streams of events~\cite{zhu2010efficient,leemans2014discovery,ao2015online}.

In this paper, we present a method to estimate the likelihood of any scenario in an uncertain setting, which covers partially ordered traces as well as other types of uncertainty illustrated in the taxonomy~\cite{DBLP:journals/informationsystems/PegoraroUA21}. Furthermore, we will cover both the non-deterministic case (\emph{strong uncertainty}) and the probabilistic case (\emph{weak uncertainty}).

\section{Running Example}\label{sec:example}
\begin{figure}[t]
	\centering
	\includegraphics[width=\columnwidth]{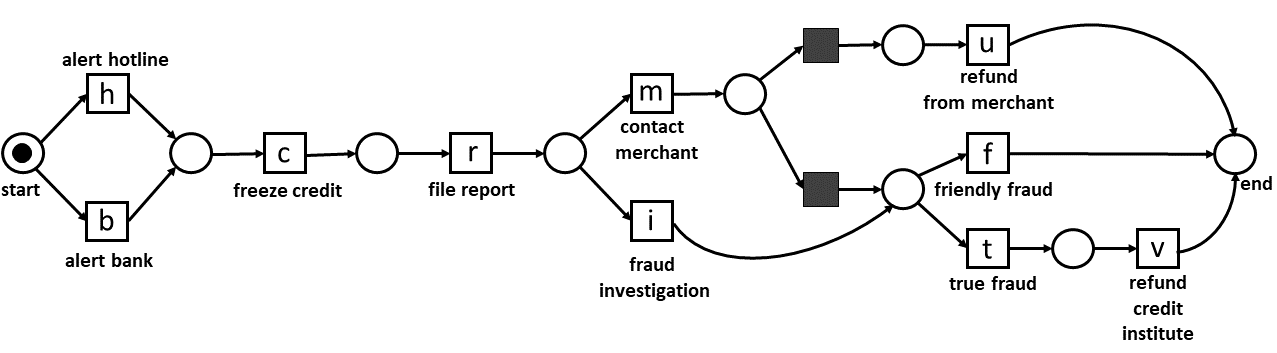}
	\caption{A Petri net model of the credit card fraud investigation process. This net allows for 10 possible traces.}
	\label{fig: petrinet}
\end{figure}

In this section, we will provide a running example of uncertain process instance related to a sample process. We will then apply our probability estimation method to this uncertain trace, to illustrate its operation.
The example we analyze here is a simplified generalization of a remote credit card fraud investigation process. This process is visualized by the Petri net in Figure~\ref{fig: petrinet}.

Firstly, the credit card owner alerts the credit card company of a possibly fraudulent transaction.
The customer may either notify the company by calling their hotline (\emph{alert hotline}) or arrange an urgent meeting with personnel of the bank that issued the credit card (\emph{alert bank}).
In both scenarios, his credit is frozen (\emph{freeze credit}) to prevent further fraud.
All information provided by the customer about the transaction is summarized when filing the formal report (\emph{file report}).
As a next step, the credit card company tries to contact the merchant that charged the credit card.
If this happens (\emph{contact merchant}), the credit card company clarifies whether there has been just a mistake (e.g., merchant charging not delivering a product, or a billing mistake) on the merchant's side.
In such cases, the customer gets a \emph{refund from merchant} and the case is closed.
Another outcome might be the discovery of a \emph{friendly fraud}, which is when a cardholder makes a purchase and then disputes it as fraud even though it was not.
If contacting the merchant is impossible, a \emph{fraud investigation} is initiated.
In this case, fraud investigators will usually start with the transaction data and look for timestamps, geolocation, IP addresses, and other elements that can be used to prove whether or not the cardholder was involved in the transaction.
The outcome might be either friendly fraud or \emph{true fraud}.
True fraud can also happen when both the merchant and the cardholder are affected by the fraud.
In this case, the cardholder receives a refund from the credit institute (activity \textit{refund credit institute}) and the case is closed.

Note that for simplicity, we have used single letters to represent the activity labels in the Petri net transitions.
Some possible traces in this process are for example:
$\langle h,c,r,m,u \rangle$,
$\langle b,c,r,m,f \rangle$,
$\langle h,c,r,i,f \rangle$ and
$\langle b,c,r,i,t,v \rangle$.

Suppose that the credit card company wants to perform conformance checking to identify deviant process instances. However, some traces in the information system of the company are affected by uncertainty, such as the one in Table~\ref{table: credit card case}.

\begin{table}[t]
	\caption{Example of an uncertain case from the credit card fraud investigation process.}
	\centering
	\scriptsize
		\begin{tabular}{ccccc}
			Case ID & Event ID & Activity & Timestamp & Ind.\\ \hline
			\multicolumn{1}{|c|}{5167} & 
			\multicolumn{1}{|c|}{$e_1$} &
			\multicolumn{1}{|c|}{$h$ (alert hotline)} &
			\multicolumn{1}{c|}{\begin{tabular}[c]{@{}c@{}} 05-10-2020 23:00
			\end{tabular}}&
			\multicolumn{1}{|c|}{} 
			\\ \hline
			\multicolumn{1}{|c|}{5167} & 
			\multicolumn{1}{|c|}{$e_2$} &
			\multicolumn{1}{|c|}{$c$ (freeze credit)} &
			\multicolumn{1}{c|}{\begin{tabular}[c]{@{}c@{}} 06-10-2020 \end{tabular}}&
			\multicolumn{1}{|c|}{} 
			\\ \hline
			\multicolumn{1}{|c|}{5167} & 
			\multicolumn{1}{|c|}{$e_3$} &
			\multicolumn{1}{|c|}{$r$ (file report)} &
			\multicolumn{1}{c|}{\begin{tabular}[c]{@{}c@{}} $U$(05-10-2020 20:00,\\ 06-10-2020 10:00) \end{tabular}}&
			\multicolumn{1}{|c|}{} 
			\\ \hline
			\multicolumn{1}{|c|}{5167} & 
			\multicolumn{1}{|c|}{$e_4$} &
			\multicolumn{1}{|c|}{$i$ (fraud investigation)} &
			\multicolumn{1}{c|}{\begin{tabular}[c]{@{}c@{}} 09-10-2020 10:00 \end{tabular}}&
			\multicolumn{1}{|c|}{} 
			\\ \hline
			\multicolumn{1}{|c|}{5167} & 
			\multicolumn{1}{|c|}{$e_5$} &
			\multicolumn{1}{|c|}{\shortstack{$\{f: 0.3$ (friendly fraud), \\ $t: 0.7$ (true fraud)$\}$}} &
			\multicolumn{1}{c|}{\begin{tabular}[c]{@{}c@{}} 14-10-2020 09:00 \end{tabular}}&
			\multicolumn{1}{|c|}{} 
			\\ \hline
			\multicolumn{1}{|c|}{5167} & 
			\multicolumn{1}{|c|}{$e_6$} &
			\multicolumn{1}{|c|}{$v$ (refund credit institute)} &
			\multicolumn{1}{c|}{\begin{tabular}[c]{@{}c@{}} 15-10-2020 10:00 \end{tabular}}&
			\multicolumn{1}{|c|}{$?$} 
			\\ \hline
		\end{tabular}
		\normalsize
		\label{table: credit card case}
%	\end{adjustbox}
\end{table} 

Suppose that in the first half of October 2020, the company was implementing a new system for automatic event data generation.
During this time, the event data regarding the credit card fraud investigation process often had to be inserted manually by the employees.
Such manual recordings were subject to inaccuracies, leading to imprecise or missing data affecting the cases during this period.
The process instance from Table~\ref{table: credit card case} is one of the affected instances.
Here, events $e_2,e_3,e_5,e_6$ are uncertain.
The timestamp of event $e_2$ is not precise enough, so the possible timestamp lies between 06-10-2020 00:00 and 06-10-2020 23:59.
Event $e_3$ has happened some time between 20:00 on October 5th and 10:00 on October 6th. 
Event $e_5$ has two possible activity labels: $f$ with probability $0.3$ and $t$ with probability $0.7$.
Refunding the customer (event $e_6$) has been recorded in the system, but the customer has not received the money yet, which is why the event is indeterminate: this is indicated with a question mark (?) in the rightmost column, and indicates an event that has been recorded, but for which is unclear if it actually occurred in reality.

The credit card company is interested in understanding if and how the data in this uncertain trace conforms with the normative process model, and the entity of the actual compliance risk; they are specifically interested in knowing whether a severely non-compliant scenario is highly likely. In the remainder of the paper, we will describe a method able to estimate the probability of all possible outcome scenarios.

\section{Preliminaries}\label{sec:prelim}
Let us now present some preliminary definitions regarding uncertain event data.

\begin{definition}[Uncertain attributes]\label{def:attr}
	Let $\mathbb{U}$ be the \emph{universe of attribute domains}, and the set $\mathcal{D} \in \mathbb{U}$ be an \emph{attribute domain}. Any $\mathcal{D} \in \mathbb{U}$ is a discrete set or a totally ordered set. A \emph{strongly uncertain attribute} of domain $\mathcal{D}$ is a subset $d_S \subseteq \mathcal{D}$ if $\mathcal{D}$ is a discrete set, and it is a closed interval $d_S = [d_{min}, d_{max}]$ with $d_{min} \in \mathcal{D}$ and $d_{max} \in \mathcal{D}$ otherwise. We denote with $S_\mathcal{D}$ the set of all such strongly uncertain attributes of domain $\mathcal{D}$. A \emph{weakly uncertain attribute} $f_\mathcal{D}$ of domain $\mathcal{D}$ is a function $f_\mathcal{D} \colon \mathcal{D} \not\to [0 ,1]$ such that $0 < \sum_{x \in \mathcal{D}}\,f_\mathcal{D}(x) \leq 1$ if $\mathcal{D}$ is finite, $0 < \int_{-\infty}^{\infty}f_\mathcal{D}(x)\,dx \leq 1$ otherwise. We denote with $W_\mathcal{D}$ the set of all such weakly uncertain attributes of domain $\mathcal{D}$. We collectively denote with $\mathcal{U}_\mathcal{D} = S_\mathcal{D} \cup W_\mathcal{D}$ the set of \emph{uncertain attributes} of domain $\mathcal{D}$.
\end{definition}

It is easy to see how a ``certain'' attribute $x$, with a value not affected by any uncertainty, can be represented through the definitions in use here: if its domain is discrete, it can be represented with the singleton $\{x\}$; otherwise, it can be represented with the degenerate interval $[x, x]$.

\begin{definition}[Uncertain events]\label{def:event}
	Let $\mathbb{U}_I$ be the \emph{universe of event identifiers}. Let $\mathbb{U}_C$ be the \emph{universe of case identifiers}. Let $A \in \mathbb{U}$ be the discrete domain of all the \emph{activity identifiers}. Let $T \in \mathbb{U}$ be the totally ordered domain of all the \emph{timestamp identifiers}. Let $O = \{?\} \in \mathbb{U}$, where the ``?'' symbol is a placeholder denoting \emph{event indeterminacy}. The \emph{universe of uncertain events} is denoted with $\mathcal{E} = \mathbb{U}_I \times \mathbb{U}_C \times \mathcal{U}_A \times \mathcal{U}_T \times \mathcal{U}_O$.
\end{definition}

The activity label, timestamp and indeterminacy attribute values of an uncertain event are drawn from $\mathcal{U}_A$, $\mathcal{U}_T$ and $\mathcal{U}_O$; in accordance with Definition~\ref{def:attr}, each of these attributes can be strongly uncertain (set of possible values or interval) or weakly uncertain (probability distribution). The indeterminacy domain is defined on a single element ``?'': thus, strongly uncertain indeterminacy may be $\{?\}$ (indeterminate event) or $\varnothing$ (no indeterminacy). In weakly uncertain indeterminacy, the ``?'' element is associated to a probability value.

\begin{definition}[Projection functions]\label{def:proj}
	For an uncertain event $e = (i, c, a, t, o) \in \mathcal{E}$, we define the following projection functions: 
	%$\pi_c(e) = c$, 
	$\pi_a(e) = a$, $\pi_t(e) = t$, $\pi_o(e) = o$.
	We define $\pi_a^{set}(e) = a$ if $a$ is strongly uncertain, and $\pi_a^{set}(e) = \{x \in \mathcal{U}_A \mid f_A(x) > 0 \}$ with $a = f_A$ otherwise.
	If the timestamp $t = [t_{min}, t_{max}]$ is strongly uncertain,
	%and $\pi_t(e) \in S_T$ with $\pi_t(e) = [t_{min}, t_{max}]$,
	we define $\pi_{t_{min}}(e) = t_{min}$ and $\pi_{t_{max}}(e) = t_{max}$. If the timestamp $t = f_T$ is weakly uncertain,
	%and $\pi_t(e) \in W_T$ with $\pi_t(e) = t_W$
	we define $\pi_{t_{min}}(e) = \operatorname*{argmin}_x (f_T(x) > 0)$ and $\pi_{t_{max}}(e) = \operatorname*{argmax}_x (f_T(x) > 0)$.
\end{definition}

\begin{definition}[Uncertain traces and logs]\label{def:trace_log}	
	$\tau \subset \mathcal{E}$ is an \emph{uncertain trace} if all the event identifiers in $\tau$ are unique and all events in $\tau$ share the same case identifier $c \in \mathbb{U}_C$. $\mathcal{T}$ denotes the universe of uncertain traces.  $L \subset \mathcal{T}$ is an \emph{uncertain log} if all the event identifiers in $L$ are unique.
\end{definition}

\begin{definition}[Realizations of uncertain traces]\label{def:real}
	Let $e, e' \in \mathcal{E}$ be two uncertain events. $\prec_\mathcal{E}$ is a strict partial order defined on the universe of strongly uncertain events $\mathcal{E}$ as $e \prec_\mathcal{E} e' \Leftrightarrow \pi_{t_{max}}(e) < \pi_{t_{min}}(e')$. Let $\tau \in \mathcal{T}$ be an uncertain trace. The sequence $\rho = \langle e_1, e_2, \dots, e_n \rangle \in \mathcal{E}^*$, with $n \leq |\tau|$, is an \emph{order-realization} of $\tau$ if there exists a total function $f \colon \{1, 2, \dots, n\} \rightarrow \tau$ such that:
	\begin{itemize}
		\item for all $1 \leq i < j \leq n$ we have that $\rho[j] \nprec_\mathcal{E} \rho[i]$,
		\item for all $e \in \tau$ with $\pi_o(e) = \varnothing$ there exists $1 \leq i \leq n$ such that $f(i) = e$.
	\end{itemize}
	We denote with $\mathcal{R}_O(\tau)$ the set of all such order-realizations of the trace $\tau$.
	
	Given an order-realization $\rho = \langle e_1, e_2, \dots, e_n \rangle \in \mathcal{R}_O(\tau)$, the sequence $\sigma \in {\mathcal{U}_A}^*$ is a \emph{realization} of $\rho$ if $\sigma \in \{\langle a_1, a_2, \dots, a_n \rangle \mid \forall_{1 \leq i \leq n} \:a_i \in \pi^{set}_a(i)\}$. We denote with $\mathcal{R}_A(\rho) \subseteq {\mathcal{U}_A}^*$ the set of all such realizations of the order-realization $\rho$. We denote with $\mathcal{R}(\tau) \subseteq {\mathcal{U}_A}^*$ the union of the realizations obtainable from all the order-realizations of $\tau$: $\mathcal{R}(\tau) = \bigcup_{\rho \in \mathcal{R}_O(\tau)} \mathcal{R}_A(\rho)$. We will say that an order-realization $\rho \in \mathcal{R}_O(\tau)$ \emph{enables} a sequence $\sigma \in {\mathcal{U}_A}^*$ if $\sigma \in \mathcal{R}_A(\rho)$.
\end{definition}

Detailing an algorithm to generate all realizations of an uncertain trace is beyond the scope of this paper. The literature illustrates a conformance checking method over uncertain data which employs a \emph{behavior net}, a Petri net able to replay all and only the realizations of an uncertain trace~\cite{pegoraro2019mining}. Exhaustively exploring all complete firing sequences of a behavior net, e.g., through its reachability graph, provides all realizations of the corresponding uncertain trace.

Given the above formalization, we can now define more clearly the research question that we are investigating in this paper. Given an uncertain trace $\tau \in \mathcal{T}$ and one of its realizations $\sigma \in \mathcal{R}(\tau)$, our goal is to obtain a procedure to reliably compute $P(\sigma \mid \tau) = ``\textit{probability of}~\sigma~\textit{given that we observe}~\tau"$.
In other words, provided that $\sigma$ corresponds to a scenario (i.e., a realization) for the uncertain trace $\tau$, we are interested in calculating the probability that $\sigma$ is the actual scenario occurred in reality, which caused the recording of the uncertain trace $\tau$ in the event log. In the next section, we will illustrate how to calculate such probabilities of uncertain traces realizations.

\section{Method}\label{sec:meth}
Before we show how we can obtain probability estimates for all realizations of an uncertain trace, it is important to state an assumption: the information on uncertainty related to a particular attribute in some event is independent of the possible values of the same attribute present in other events, and it is independent of the uncertainty information on other attributes of the same event. Note that in the examples of uncertainty sources given in Section~\ref{sec:introduction} (data coarseness and sensor errors), this independence assumption often holds.

Additionally, we need to consider the fact that strongly uncertain attributes do not come with known probability values: their description only specifies the values that attributes might acquire, but not the likelihood of each possible value. As a consequence, estimating probability for specific realizations in a strongly uncertain environment is only possible with a-priori assumptions on how probability distributes among the attribute value. At times, it might be possible to assume the distribution in an informed way---for instance, on the basis of features of the information system hosting the data, of the sensors recording events and attributes, or other tools involved in the management of the process.

In case no indication is present, a reasonable assumption---which we will hold for the remainder of the paper---is that any possible value of a strongly uncertain attribute is equally likely.
Formally, with $e = (i, c, a, t, o) \in \mathcal{E}$ let $\tau_s \colon \mathcal{E} \to \mathcal{E}$ be a function such that $\tau_s(e) = (i, c, a', t', o')$, where $a' = \{(x, \frac{1}{|\pi_a^{set}(e)|}) \mid x \in \pi_a^{set}(e)\}$ if $a \in S_A$ and $a' = a $ otherwise; $t' = U(\pi_{t_{min}}(e), \pi_{t_{max}}(e))$ if $t \in S_T$ and $t' = t$ otherwise; $o' = 0.5$ if $o = \{?\}$ and $o' = o$ otherwise.

First, observe that the probability $P(\sigma \mid \tau)$ that an activity sequence $\sigma \in {\mathcal{U}_A}^*$ is indeed a realization of the trace $\tau \in \mathcal{T}$, and thus $\sigma \in \mathcal{R}(\tau)$, increases with the number of order-realizations enabling it.
Furthermore, for each such order-realizations, one can construct a probability function $P_O(\rho \mid \tau)$ reflecting the likelihood of the sequence $\rho$ itself given the trace $\tau$, and a probability function $P_A(\sigma \mid \rho)$ reflecting the likelihood that the realization corresponding to $\rho$ is indeed $\sigma$.
The value of $P_O(\rho \mid \tau)$ is affected by the uncertainty information in timestamps and indeterminate events, while the value of $P_A(\sigma \mid \rho)$ is aggregated from the uncertainty information in the activity labels.

Given a realization $\sigma$ of an uncertain process instance and the set of its enablers, its probability is computed as following:

\scriptsize
\begin{align*}
P(\sigma \mid \tau) = \sum_{\rho \in \mathcal{E}^*}P_O(\rho \mid \tau) \cdot P_A(\sigma \mid \rho)
\end{align*}
\normalsize 

Note that, if $\rho$ does not enable $\sigma$, $P_A(\sigma \mid \rho) = 0$. For any uncertain trace $\tau \in \mathcal{T}$, it holds that $\sum_{\sigma \in \mathcal{R}(\tau)} P(\sigma \mid \tau) = 1$, since both $P_O(\cdot)$ and $P_A(\cdot)$ are each constructed to be (independent) probability distributions.

We will now compute $P_A(\sigma \mid \rho)$ using the information on the activity labels uncertainty.
Let us write $f_A^e$ as a shorthand for $\pi_a(e)$. If there is uncertainty in activities, then for each event $e \in \rho$ and activity label $a \in \pi_a^{set}(e)$, the probability that $e$ executes $a$ is given by $f_A^e(a)$.
Thus, for every $\rho = \langle e_1,...,e_n \rangle \in \mathcal{R}_O(\tau)$ and $\sigma = \langle a_1,...,a_n \rangle \in \mathcal{R}_O(\tau)$, the value $P_A$ can be aggregated from these distributions in the following way:

\scriptsize
\begin{align*}
P_A(\sigma \mid \rho) = \prod_{i=1}^{n} f_A^{i}(a_i)
\end{align*}
\normalsize

Through the value of $P_A$, we can assess the likelihood that any given order-realization executes a particular realization.
The next step is to estimate the probability of each order-realization $\rho$ from the set $\mathcal{R}_O(\tau)$. The probability of observing $\rho$ needs to be aggregated from the probability that the corresponding set of events appears in the given particular order, which is determined by the timestamp intervals and, if applicable, the distributions over them; and the probability that the order-realization contains the corresponding specific set of events, which is determined by the uncertainty information on the indeterminacy. Multiplying the two values obtained above to yield a probability estimate for the order-realization reflects our independence assumption. Let us firstly focus on uncertainty on timestamps, which causes the events to be partially ordered.

We will write $f_T^e(t)$ as a shorthand for $\pi_t(e)(t)$. For every event $e$, the value of $f_T^e(t)$ yields the probability that event $e$ happened on timestamp $t$.
This value is always 0 for all $t < \pi_{t_{min}}(e)$ and $t > \pi_{t_{max}}(e)$ (see $\pi_{t_{min}}$ and $\pi_{t_{max}}$ in Definition~\ref{def:proj}).
Given the continuous domain of timestamps, $P_O(\cdot)$ is assessed by using integrals.
For a trace $\tau \in \mathcal{T}$ and an order-realization $\rho = \langle e_1,...,e_n \rangle \in \mathcal{R}_O(\tau)$, let $a_i = \pi_{t_{min}}(i)$ and $b_i = \pi_{t_{max}}(i)$ for all $1 \leq i \leq n$.
Then, we define:

\scriptsize
\begin{align*}
I(\rho) &= \int_{a_1}^{min\{b_1, \dots, b_n\}} f_T^{e_1}(x_1)
\int_{max\{a_2, x_1\}}^{min\{b_2, \dots, b_n\}} f_T^{e_2}(x_2)
\cdots \\
&
\int_{max\{a_i, x_{i-1}\}}^{min\{b_i,\dots,b_n\}} f_T^{i}(x_i)
\,\cdots 
\int_{max\{a_n, x_{n-1}\}}^{b_n} f_T^{e_n}(x_n) \,
d_{x_n} \dots d_{x_1} \\
&=
\int_{a_1}^{min\{b_1, \dots, b_n\}} 
\int_{max\{a_2, x_1\}}^{min\{b_2, \dots, b_n\}} 
\cdots
\int_{max\{a_i, x_{i-1}\}}^{min\{b_i, \dots, b_n\}} \cdots \int_{max\{a_n, x_{n-1}\}}^{b_n} \:
\prod_{i=1}^{n} f_T^{i}(x_i) \,
d_{x_n} \dots d_{x_1}
\end{align*}
\normalsize

This chain of integrals allows us to compute the probability of a specific order among all the events in an uncertain trace.
Now, to compute the probability of each realization from $\mathcal{R}_e$ accounting for indeterminate events, we combine both the probability of the events having appeared in a particular order and the probability that the sequence contains exactly those events.
For simplicity, we will use a function that acquires the value 1 if an event is not indeterminate. Let us define $f_O^e \colon O \to [0, 1]$ such that $f_O^e(?) = \pi_o(e)(?)$ if $\pi_o(e) \neq \varnothing$ and $f_O^e(?) = 1$ otherwise.
More precisely, given $\tau \in \mathcal{T}$ and $\rho \in \mathcal{R}_O(\tau)$, we compute:

\scriptsize
\begin{align*}
P_O(\rho \mid \tau)= I(\rho) \cdot \prod_{\substack{e \in \tau \\ e \in \rho}} (1 - f_O^{e}(?)) \cdot
\prod_{\substack{e \in \tau \\ e \not \in \rho}} f_O^{e}(?)
\end{align*}
\normalsize

We now have at our disposal all the necessary tools to compute a probability distribution over the trace realizations of any uncertain process instance in any possible uncertainty scenario. Let us then apply this method to compute the probabilities of all realizations of the trace $\tau$ in Table~\ref{table: credit card case}, and to analyze its conformance to the process in Figure~\ref{fig: petrinet}.

Each order-realization of $\tau$ enables two realizations, because event $e_5$ has two possible activity labels.
Since for events $e \in \tau \setminus \{e_5\}$, we have $f_A^e$ equal to 1 for their corresponding unique activity label, the probability that an order-realization $\rho \in \mathcal{R}_O(\tau)$ has some realization $\sigma \in \mathcal{R}_A(\rho)$ only depends on whether the trace $\sigma$ contains activity $f$ or $t$.
Thus, for traces $\sigma^{1'},\sigma^{2'},\sigma^{3'},\sigma^{4'},\sigma^{5'},\sigma^{6'}$ and their unique enabling sequences, we always have \mbox{$P_A(\sigma^{i'} \mid s_e^{i})$} $= f_A^{e_5}(f) = 0.3$, where $i \in \{1, \dots, 6\}$.
Similarly, for traces $\sigma^{1''},\sigma^{2''},\sigma^{3''},\sigma^{4''},\sigma^{5''},\sigma^{6''}$ and their unique enabling sequences, we always have \mbox{$P_A(\sigma^{i''} \mid \rho^{i})$} $= f_A^{e_5}(t) = 0.7$, where \mbox{$i \in \{1, \dots, 6\}$}. Next, we calculate the $P_O(\cdot)$ values for the 6 possible order-realizations in $\mathcal{R}_O(\tau)$, which are displayed in Table~\ref{table:o-real}.

\begin{table}[t]
	\centering
	\begin{minipage}[]{.43\textwidth}
		\caption{The possible order-realizations of the process instance from Table~\ref{table: credit card case} and their probabilities.}
		\centering
		\scriptsize
		\begin{adjustbox}{width=\columnwidth, center}
			\begin{tabular}{ccc}
				Order-realization $\rho$ & \textbf{$I(\rho)$} & \textbf{$P_O(\rho)$}
				\\ \hline
				\multicolumn{1}{|l|}{$\rho^1{:}\langle e_1, e_2, e_3, e_4, e_5, e_6 \rangle$} & 
				\multicolumn{1}{|c|}{$0.140$} &
				\multicolumn{1}{|c|}{$0.074$} 
				\\ \hline
				\multicolumn{1}{|l|}{$\rho^2{:}\langle e_1, e_3, e_2, e_4, e_5, e_6\rangle$} & 
				\multicolumn{1}{|c|}{$0.780$} &
				\multicolumn{1}{|c|}{$0.390$} 
				\\ \hline
				\multicolumn{1}{|l|}{$\rho^3{:}\langle e_3,e_1,e_2,e_4,e_5,e_6\rangle$} & 
				\multicolumn{1}{|c|}{$0.072$} &
				\multicolumn{1}{|c|}{$0.036$} 
				\\ \hline
				\multicolumn{1}{|l|}{$\rho^4{:}\langle e_1,e_2,e_3,e_4,e_5\rangle$} & 
				\multicolumn{1}{|c|}{$0.149$} &
				\multicolumn{1}{|c|}{$0.074$} 
				\\ \hline
				\multicolumn{1}{|l|}{$\rho^5{:}\langle e_1,e_3,e_2,e_4,e_5\rangle$} & 
				\multicolumn{1}{|c|}{$0.780$} &
				\multicolumn{1}{|c|}{$0.390$} 
				\\ \hline
				\multicolumn{1}{|l|}{$\rho^6{:}\langle e_3,e_1,e_2,e_4,e_5\rangle$} & 
				\multicolumn{1}{|c|}{$0.072$} &
				\multicolumn{1}{|c|}{$0.036$} 
				\\ \hline
			\end{tabular}
		\end{adjustbox}
		\normalsize
		\label{table:o-real}
	\end{minipage}%
	\quad
	\begin{minipage}[]{.53\textwidth}
		\caption{The set of possible realizations of the example from Table~\ref{table: credit card case}, their enablers, their probabilities, and their conformance scores. The conformance score is equal to the cost of the optimal alignment between the trace and the Petri net in Figure~\ref{fig: petrinet}.}
		\centering
		\scriptsize
		\begin{adjustbox}{width=\columnwidth, center}
			\begin{tabular}{cccc}
				Realization $\sigma$ & $\rho$ & $P(\sigma \mid \tau)$ & $conf$
				\\ \hline
				\multicolumn{1}{|l|}{$\sigma^{1'}{:}\langle h,c,r,i,f,v \rangle$} & 
				\multicolumn{1}{|c|}{$\rho^1$} & 
				\multicolumn{1}{|c|}{$P_O(\rho^1) {\cdot} P_A(\sigma^{1'} | \rho^1) = 0.022$} &
				\multicolumn{1}{|c|}{$1$}
				\\ \hline
				\multicolumn{1}{|l|}{$\sigma^{1''}{:}\langle h,c,r,i,t,v \rangle$} & 
				\multicolumn{1}{|c|}{$\rho^1$} & 
				\multicolumn{1}{|c|}{$P_O(\rho^1) {\cdot} P_A(\sigma^{1''} | \rho^1) = 0.052$} &
				\multicolumn{1}{|c|}{$0$}
				\\ \hline
				\multicolumn{1}{|l|}{$\sigma^{2'}{:}\langle h,r,c,i,f,v\rangle$} & 
				\multicolumn{1}{|c|}{$\rho^2$} & 
				\multicolumn{1}{|c|}{$P_O(\rho^2) {\cdot} P_A(\sigma^{2'} | \rho^2) =  0.117$} &
				\multicolumn{1}{|c|}{$3$}
				\\ \hline
				\multicolumn{1}{|l|}{$\sigma^{2''}{:}\langle h,r,c,i,t,v\rangle$} & 
				\multicolumn{1}{|c|}{$\rho^2$} & 
				\multicolumn{1}{|c|}{$P_O(\rho^2) {\cdot} P_A(\sigma^{2''} | \rho^2) = 0.273$} &
				\multicolumn{1}{|c|}{$2$}
				\\ \hline
				\multicolumn{1}{|l|}{$\sigma^{3'}{:}\langle r,h,c,i,f,v\rangle$} & 
				\multicolumn{1}{|c|}{$\rho^3$} & 
				\multicolumn{1}{|c|}{$P_O(\rho^3) {\cdot} P_A(\sigma^{3'} | \rho^3) = 0.011$} &
				\multicolumn{1}{|c|}{$3$}
				\\ \hline
				\multicolumn{1}{|l|}{$\sigma^{3''}{:}\langle r,h,c,i,t,v\rangle$} & 
				\multicolumn{1}{|c|}{$\rho^3$} & 
				\multicolumn{1}{|c|}{$P_O(\rho^3) {\cdot} P_A(\sigma^{3''} | \rho^3) = 0.025$} &
				\multicolumn{1}{|c|}{$2$}
				\\ \hline
				\multicolumn{1}{|l|}{$\sigma^{4'}{:}\langle h,c,r,i,f \rangle$} & 
				\multicolumn{1}{|c|}{$\rho^4$} & 
				\multicolumn{1}{|c|}{$P_O(\rho^4) {\cdot} P_A(\sigma^{4'} | \rho^4) = 0.022$} &
				\multicolumn{1}{|c|}{$0$}
				\\ \hline
				\multicolumn{1}{|l|}{$\sigma^{4''}{:}\langle h,c,r,i,t \rangle$} & 
				\multicolumn{1}{|c|}{$\rho^4$} & 
				\multicolumn{1}{|c|}{$P_O(\rho^4) {\cdot} P_A(\sigma^{4''} | \rho^4) = 0.052$} &
				\multicolumn{1}{|c|}{$1$}
				\\ \hline
				\multicolumn{1}{|l|}{$\sigma^{5'}{:}\langle h,r,c,i,f\rangle$} & 
				\multicolumn{1}{|c|}{$\rho^5$} & 
				\multicolumn{1}{|c|}{$P_O(\rho^5) {\cdot} P_A(\sigma^{5'} | \rho^5) = 0.117$} &
				\multicolumn{1}{|c|}{$2$}
				\\ \hline
				\multicolumn{1}{|l|}{$\sigma^{5''}{:}\langle h,r,c,i,t\rangle$} & 
				\multicolumn{1}{|c|}{$\rho^5$} & 
				\multicolumn{1}{|c|}{$P_O(\rho^5) {\cdot} P_A(\sigma^{5''} | \rho^5) = 0.273$} &
				\multicolumn{1}{|c|}{$3$}
				\\ \hline
				\multicolumn{1}{|l|}{$\sigma^{6'}{:}\langle r,h,c,i,f\rangle$} & 
				\multicolumn{1}{|c|}{$\rho^6$} & 
				\multicolumn{1}{|c|}{$P_O(\rho^6) {\cdot} P_A(\sigma^{6'} | \rho^6) = 0.011$} &
				\multicolumn{1}{|c|}{$2$}
				\\ \hline
				\multicolumn{1}{|l|}{$\sigma^{6''}{:}\langle r,h,c,i,t\rangle$} & 
				\multicolumn{1}{|c|}{$\rho^6$} & 
				\multicolumn{1}{|c|}{$P_O(\rho^6) {\cdot} P_A(\sigma^{6''} | \rho^6) = 0.025$} &
				\multicolumn{1}{|c|}{$3$}
				\\ \hline
			\end{tabular}
		\end{adjustbox}
		\normalsize
		\label{table: p values}
	\end{minipage}
\end{table}

One can notice that the $I$ values only depend on the ordering of the first three events, which are also the only ones with overlapping timestamps.
Since the indeterminate event $e_6$ does not overlap with any other event, pairs of sequences where the first three events have the same order also have the same probability.
This reflects our assumption that the occurrence and non-occurrence of $e_6$ are both equally possible.
Table \ref{table: p values} displays the calculations for the computation of the $P(\sigma \mid \tau)$ values for all realizations.
Now we can compute the expected conformance score for the uncertain process instance $\tau = \{e_1, \dots, e_6\}$. We can do so by computing alignments~\cite{van2017aligning} for each realization of $\tau$:

\scriptsize
\begin{align*}
\overline{conf}(\tau) &= \sum_{\sigma \in \mathcal{R}(\tau)} P(\sigma \mid \tau) \cdot conf(\sigma,M) = 0.022 \cdot 1 + 0.05 \cdot 0 + 0.117 \cdot 3 + 0.273 \cdot 2 + 0.011 \cdot 3 \\ 
&+ 0.025 \cdot 2 +
0.022 \cdot 0 + 0.052 \cdot 1 + 0.117 \cdot 2 + 0.273 \cdot 3 + 0.011 \cdot 2 + 0.025 \cdot 3 \\
&= 2.204.
\end{align*}
\normalsize

Given the information on uncertainty available for the trace, this conformance score is a more realistic estimate of the real conformance score compared to taking the best, worst or average scores with values 0, 3 and 1.75 respectively.

\section{Validation of Probability Estimates}\label{sec:exp}
\begin{figure}[t]
	\centering
	\begin{minipage}[t]{.43\textwidth}
		\centering
		
		\begin{tikzpicture}[->,>=stealth',shorten >=1pt,node 						distance=2.5cm,auto,main node/.style={circle,draw,align=center}, scale=.6]
		%\draw [help lines] (0,0) grid (8,5);
		\node[main node,label=above:  \small{$a$}] (A) at (2,2) {$e_1$};
		\node[main node,label=above: \large{$\substack{b:~0.9\\ c:~0.1}$}] (B) at (4,3) {$e_2$};
		\node[main node,dashed,label=above: \small{$d$}, label=below:  ?:~0.8] (C) at (4,1) {$e_3$};
		\node[main node,label=above:  \small{$e$}] (D) at (6,2) {$e_4$};

		\path
		(A) edge (B)
		(A) edge (C)
		(C) edge (D)
		(B) edge (D)

		;
		\end{tikzpicture}
		\captionof{figure}{The behavior graph of the uncertain trace considered as example for validation. %with strong uncertainty in timestamps and weak uncertainty in activities and indeterminacy.
		}
		\label{fig: validation beh graph}
	\end{minipage}
	\quad
	\begin{minipage}[t]{.53\textwidth}
		\centering
		\includegraphics[width=\linewidth, keepaspectratio]{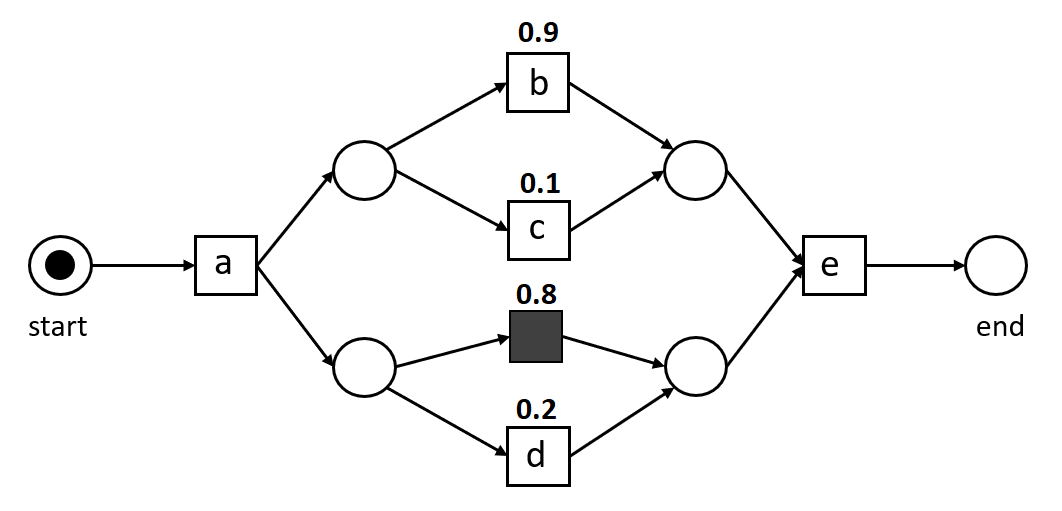}
		\captionsetup{width=.9\linewidth}
		\caption{The behavior net obtained from the behavior graph in Figure~\ref{fig: validation beh graph}.}
		\label{fig: behavior net}
	\end{minipage}
\end{figure}

In this section, we compute the probability estimates for the realizations of an uncertain trace, and then show a validation of those estimates by Monte Carlo simulation on the behavior net of the trace. The process instance of our example has strong uncertainty in timestamps and weak uncertainty in activities and indeterminacy.
It consists of 4 events: $e_1, e_2, e_3$ and $e_4$, where $e_2$ and $e_3$ have overlapping timestamps.
Event $e_2$ executes $b$ (resp., $c$) with probability 0.9 (resp., 0.1).
There is a probability of 0.2 that $e_3$ did not occur.
Figure~\ref{fig: validation beh graph} shows the corresponding behavior graph, an uncertain event data visualization that represents the time relationships between events with a directed acyclic graph~\cite{pegoraro2019mining}.
Lastly, Table~\ref{table: validation estimates} list all the possible realizations, their probabilities, and the order-realizations enabling them.

\begin{table}[t]
	\centering
	\scriptsize
	\caption{The set of realizations of the trace from Figure~\ref{fig: validation beh graph}, their enablers, and their probabilities.}
	\begin{tabular}{ccc}
		Realization $\sigma$ & $\rho$ & \textbf{$P(\sigma | \tau)$}
		\\ \hline
		\multicolumn{1}{|l|}{$\sigma^{1}{:} \langle a,b,e \rangle$} &                                                 	\multicolumn{1}{|l|}{$\rho^{1}{:} \langle e_1,e_2,e_4 \rangle$} & 
		\multicolumn{1}{|l|}{$P_O(\rho^1) {\cdot} P_A(\sigma^1 | \rho^1) 
			=  0.8 {\cdot} 0.9  = 0.72$}
		\\ \hline  
		\multicolumn{1}{|l|}{$\sigma^{2}{:} \langle a,b,d,e \rangle$} & 
		\multicolumn{1}{|l|}{$\rho^2{:} \langle e_1,e_2,e_3,e_4 \rangle$} & 
		\multicolumn{1}{|l|}{$P_O(\rho^2) {\cdot} P_A(\sigma^2 | \rho^2) = (0.5 {\cdot} 0.2) {\cdot} 0.9 = 0.09$}
		\\ \hline
		\multicolumn{1}{|l|}{$\sigma^{3}{:} \langle a,d,b,e \rangle$} & 
		\multicolumn{1}{|l|}{$\rho^3{:} \langle e_1,e_3,e_2,e_4 \rangle$} & 
		\multicolumn{1}{|l|}{$P_O(\rho^3) {\cdot} P_A(\sigma^3 | \rho^3) =  (0.5 {\cdot} 0.2) {\cdot} 0.9 = 0.09$}
		\\ \hline
		\multicolumn{1}{|l|}{$\sigma^{4}{:} \langle a,c,e \rangle$} &                                                       	\multicolumn{1}{|l|}{$\rho^{4}{:} \langle e_1,e_2,e_4 \rangle$} & 
		\multicolumn{1}{|l|}{$P_O(\rho^4) {\cdot} P_A(\sigma^4 | \rho^4)
			=0.8 {\cdot} 0.1 = 0.08$ }
		\\ \hline  
		\multicolumn{1}{|l|}{$\sigma^{5}{:} \langle a,c,d,e \rangle$} & 
		\multicolumn{1}{|l|}{$\rho^5{:} \langle e_1,e_2,e_3,e_4 \rangle$} & 
		\multicolumn{1}{|l|}{$P_O(\rho^5) {\cdot} P_A(\sigma^5 | \rho^5) = (0.5 {\cdot} 0.2) {\cdot} 0.1 = 0.01$}
		\\ \hline
		\multicolumn{1}{|l|}{$\sigma^{6}{:} \langle a,d,c,e \rangle$} & 
		\multicolumn{1}{|l|}{$\rho^6{:} \langle e_1,e_3,e_2,e_4 \rangle$} & 
		\multicolumn{1}{|l|}{$P_O(\rho^6) {\cdot} P_A(\sigma^6 | \rho^6) = (0.5 {\cdot} 0.2) {\cdot} 0.1 = 0.01$}
		\\ \hline
	\end{tabular}
	\normalsize
	\label{table: validation estimates}
\end{table}

We now validate our obtained probability estimates quantitatively by means of a Monte Carlo simulation approach.
First, we construct the behavior net~\cite{DBLP:journals/informationsystems/PegoraroUA21} corresponding to the uncertain process instance, which is shown in Figure~\ref{fig: behavior net}.
The set of replayable traces in this behavior net is exactly the set of realizations for the uncertain instance.
Then, we simulate realizations on the behavior net, dividing the accumulated count of each realization by the number of runs, and compare those values to our probability estimates.
Here, we use the \emph{stochastic simulator} of the PM4Py library~\cite{berti2019process}.
In every step of the simulation, the stochastic simulator chooses one enabled transition to fire according to a stochastic map, assigning a weight to each transition in the Petri net (here, the behavior net). 
%The following transition to fire is chosen randomly with a probability distribution, computed over the set of enabled transitions by normalizing their weights.

To simulate uncertainty in activities, events and timestamps, we do the following: 
possible activities executed by the same event appearing in an XOR-split in the behavior net are weighted so to reflect the probability values of the activity labels. 
Indeterminacy is equivalently modeled as an XOR-choice between a visible transition and a silent one in the behavior net, so to model a ``skip''. 
If there are two or more possible activities for an indeterminate event, then the sum of the weights of the visible transitions in relation to the weight of the silent transition should be the same as in the distribution given in the event type uncertainty information.
%Regarding timestamps, through the simulator we can only simulate strong uncertainty, that is, uniform distribution over the possible event orderings. 
Whenever there are events with overlapping timestamps, these appear in an AND-split in the behavior net. 
The (enabled) path of the AND-split which is taken first signals which event is executed at that moment.
%Since one event might have many possible activities and thus transitions, to simulate a uniform distribution over the choice of the path at any moment, the sum of transitions' weights enabled immediately after following each path should sum to 1.

Let $bn(\tau) = (P, T)$ be the behavior net of trace $\tau$.
Let $(e,a) \in T$ be a visible transition related to some event $e \in \tau$.
We weight $(e,a)$ the following way:

\scriptsize
\begin{align*}
weight((e,a))  = \begin{cases}
f_A^{e}(a) & \mbox{if} \; \pi_o(e) = \varnothing,  \\
(1 - f_O^e(?)) \cdot f_A^{e}(a) & \mbox{otherwise}.
\end{cases} 
\end{align*}
\normalsize

If $e \in \tau$ is an indeterminate event, then $weight((e,\epsilon)) = f_O^{e}(?)$.

Note that according to the weight assignment function, if $e$ is determinate, then $\sum_{a \in \pi_a^{set}(e)} \allowbreak weight((e,a)) = 1$.
Otherwise, $\sum_{a \in \pi_a^{set}(e)} weight((e,a)) = 1 - f_O^{e}(?) = 1 - weight((e,\tau))$.
By construction of the behavior net, any transition related to an event in $\tau$ can only fire in accordance with the partial order of uncertain timestamps.
Additionally, all transitions representing events with overlapping timestamps appear in an AND construct.
By definition of our weight function, whenever the transitions of some $e \in \tau$ are enabled (in an XOR construct), the probability of firing one of them is $1/k$, where $k$ is the number of events from $\tau$ for which none of the corresponding transitions have fired yet.
This way, there is always a uniform distribution over the set of enabled transitions representing overlapping events. Assigning the weights according to this distribution allows to decorate the behavior net with probabilities that reflect the chances of occurrence of every possible value in uncertain attributes.

\begin{figure}[t]
	\centering
	\begin{minipage}[t]{.48\textwidth}
		\centering
		\includegraphics[width=\linewidth, keepaspectratio]{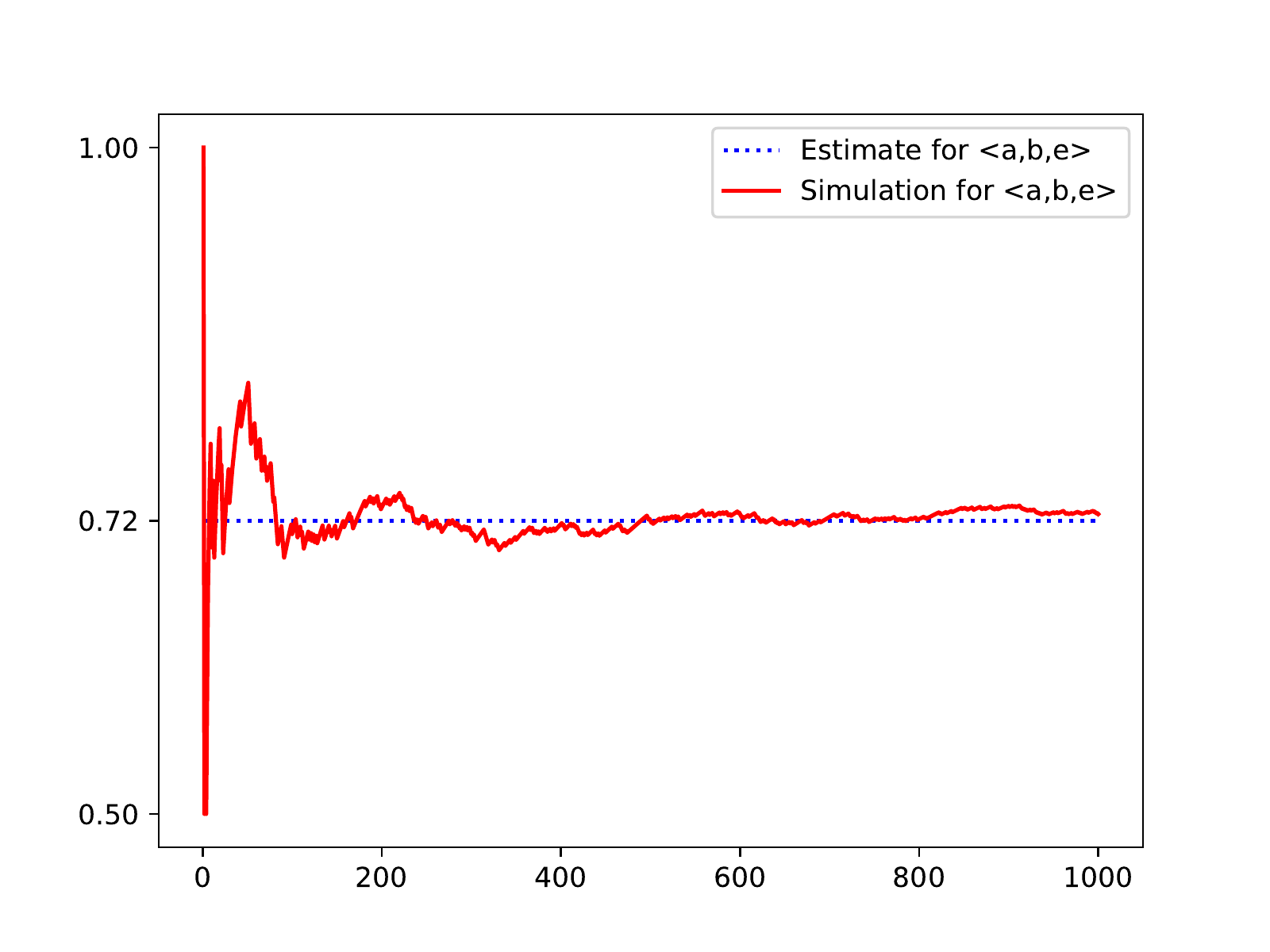}
		\caption{Plot showing how the frequency of trace $\langle a,b,e \rangle$ converges to the expected value of $0.72$ over 1000 runs.}
		\label{fig: abe}
	\end{minipage}
	\quad
	\begin{minipage}[t]{.48\textwidth}
		\centering
		\includegraphics[width=\linewidth, keepaspectratio]{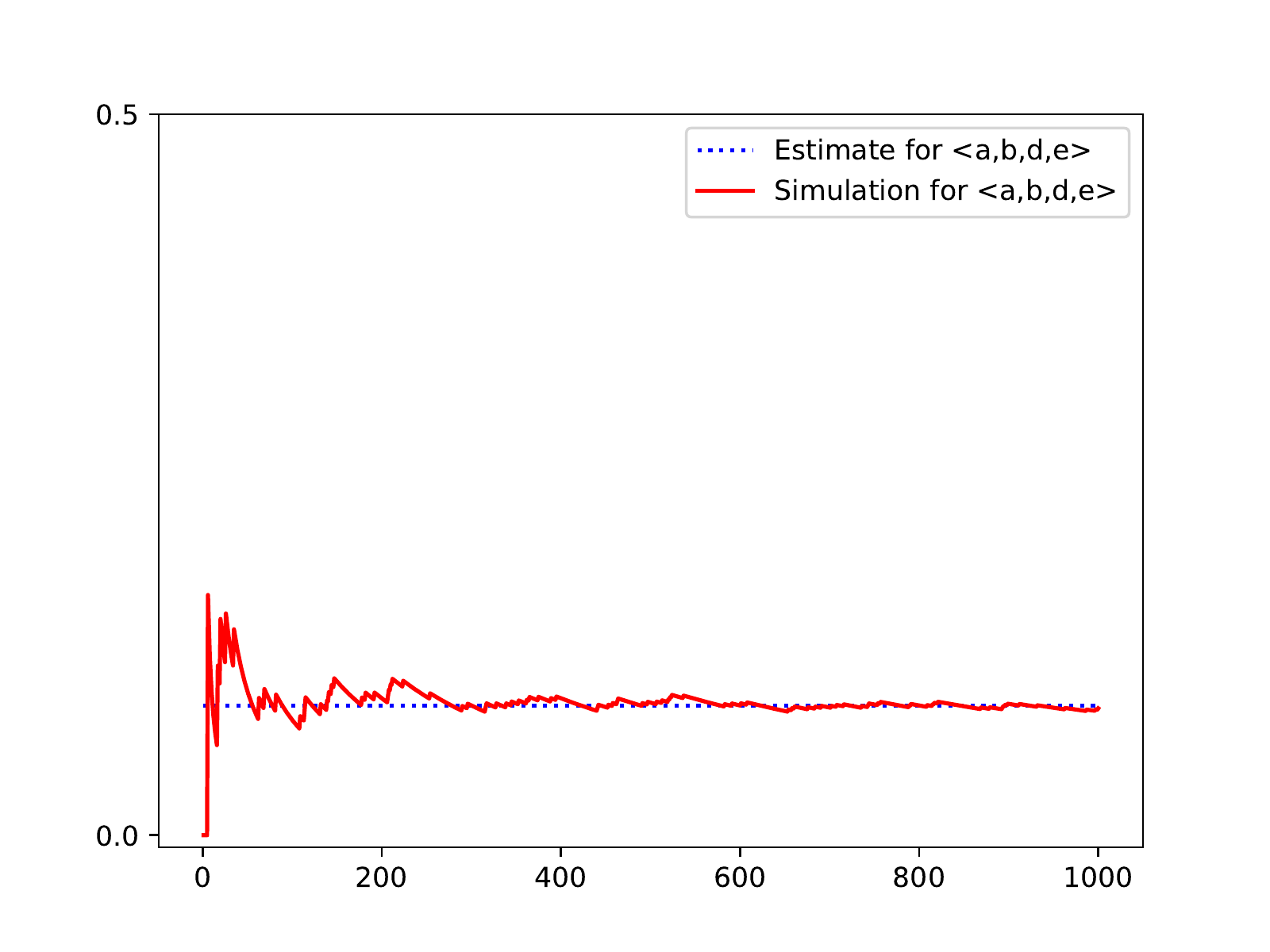}
		\caption{Plot showing how the frequency of trace $\langle a,b,d,e \rangle$ converges to the expected value of $0.09$ over 1000 runs.}%
		\label{fig: abde}
	\end{minipage}
\end{figure}

\begin{figure}[t]
	\centering
	\begin{minipage}[t]{.48\textwidth}
		\centering
		\includegraphics[width=\linewidth, keepaspectratio]{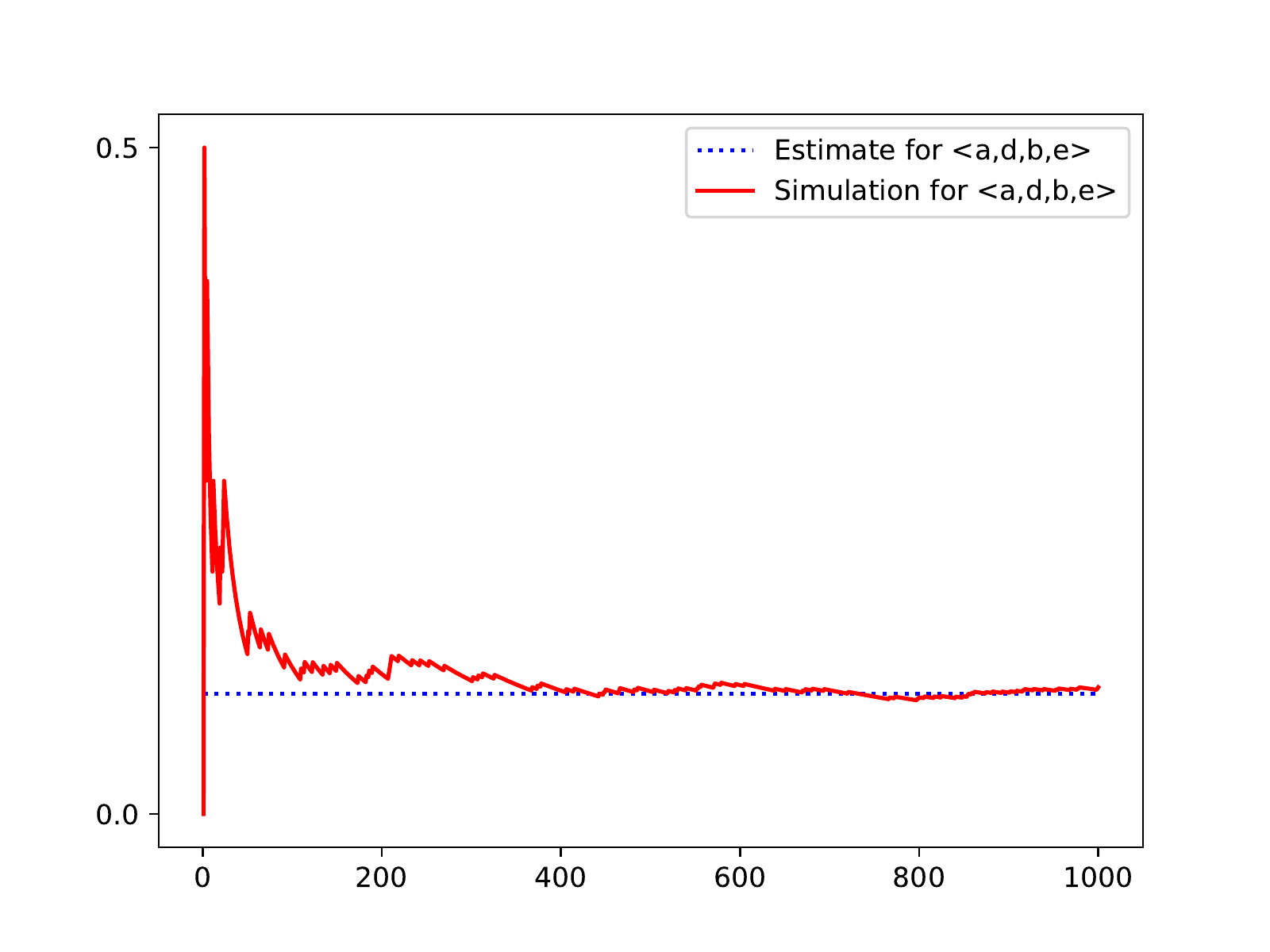}
		\caption{Plot showing how the frequency of trace $\langle a,d,b,e \rangle$ converges to the expected value of $0.09$ over 1000 runs.}
		\label{fig: adbe}
	\end{minipage}
	\quad
	\begin{minipage}[t]{.48\textwidth}
		\centering
		\includegraphics[width=\linewidth, keepaspectratio]{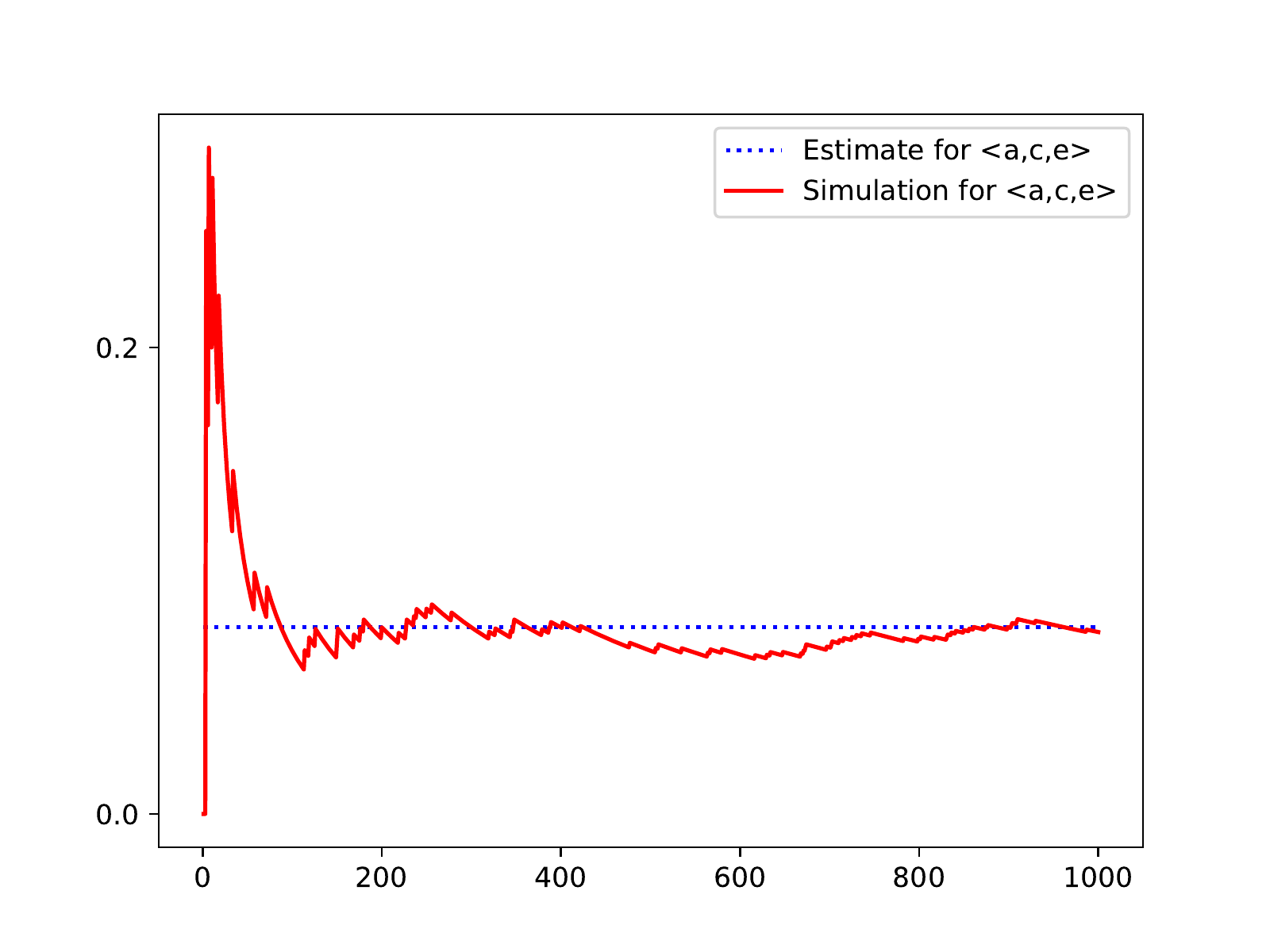}
		\caption{Plot showing how the frequency of trace $\langle a,c,e \rangle$ converges to the expected value of $0.08$ over 1000 runs.}
		\label{fig: ace}
	\end{minipage}
\end{figure}

Applying the stochastic simulator $n$ times yields $n$ realizations.
For each of the 6 possible realizations for the uncertain process instance, we obtain a probability measurement by dividing its simulated frequency by $n$.
Figures~\ref{fig: abe} through~\ref{fig: ace} show how for greater $n$, this measurement converges to the probability estimates shown in Table~\ref{table: validation estimates}, which were computed with our method.

To conclude, the Monte Carlo simulation shows that our estimated probabilities for realizations match their relative frequencies when one simulates the behavior net of the corresponding uncertain trace.

\section{Conclusion}\label{sec:conc}
Uncertain traces inherently contain behavior, allowing for many realizations; these, in turn, correspond to diverse possible real-life scenarios, that may have different consequences on the management and governance of a process. In this paper, we presented a method to quantify the probability of each realization of an uncertain trace. This enables process analysts to weigh the impact of specific insights gathered with uncertainty-aware process mining techniques, such as conformance checking using alignments. As a consequence, information from process analysis techniques can be associated with a quantification of risk or opportunity for specific scenarios, making them more trustworthy.

Multiple avenues for future work on this topic are possible. These include inferring probabilities for uncertain traces from sections of the log not affected by uncertainty, adopting certain traces or fragments of traces as ground truth. Moreover, inferring probabilities by examining evidence against a ground truth can also be achieved with a normative model that includes information concerning the probability of error or noise in specific parts of the process.

\bibliographystyle{splncs04}
\bibliography{bibliography}

\end{document}